\PassOptionsToPackage{frozencache}{markdown}
\documentclass[10pt, a4paper, oneside]{article}
\usepackage[hidelinks]{hyperref}
\usepackage{jucs2e}
\usepackage{graphicx}
\usepackage{url}
\usepackage{ulem}
\usepackage{mathtools}
\usepackage{scalerel}
\usepackage{setspace}
\usepackage[strict]{changepage}
\usepackage{caption}
\usepackage[letterspace=0]{microtype}
\usepackage{newtxtext}
\usepackage{afterpage}
\usepackage{ragged2e}
\usepackage{titlesec}

\titleformat*{\section}{\Large\bfseries}
\titleformat*{\subsection}{\normalsize\bfseries}

\graphicspath{{./figure/}}

\usepackage[textwidth=8cm, margin=0cm, left=4.6cm, right=4.2cm, top=3.9cm, bottom=6.8cm, a4paper, headheight=0.5cm, headsep=0.5cm]{geometry}
\usepackage{fancyhdr}
\usepackage[format=plain, labelfont=it, textfont=it, justification=centering]{caption}
\usepackage{breakcites}
\usepackage{microtype}
 
\apptocmd{\frame}{}{\justifying}{}

\newcommand\jucs{{Journal of Universal Computer Science}}
\newcommand\jucsvol{vol. 28, no. 2 (2022)}
\newcommand\jucspages{181-201}
\newcommand\jucssubmitted{2/6/2021}
\newcommand\jucsaccepted{4/1/2022}
\newcommand\jucsappeared{28/2/2022}
\newcommand\jucslicence{ CC BY-ND 4.0}
\newcommand\startingPage{181}
\newcommand\paperauthor{{Novotný V., Štefánik M., Ayetiran E.F., Sojka P., Řehůřek R.: }}

\newcommand\papertitle{When FastText Pays \ldots}
\newcommand\longpapertitle{When FastText Pays Attention: \\ Efficient Estimation of Word Representations \\ using Constrained Positional Weighting}

\header{\paperauthor \papertitle}

\usepackage{amsmath,amsfonts,amssymb,nicefrac,latexsym}
\usepackage{multirow}
\usepackage{url}
\usepackage[all]{nowidow}
\usepackage[
  style=alphabetic,
  firstinits=true,
  maxbibnames=100,
]{biblatex}
\defbibenvironment{bibliography}
 {\list
     {\printtext[labelalphawidth]{%
        \printfield{prefixnumber}%
        \printfield{labelalpha}%
        \printfield{extraalpha}}}
 {\setlength{\labelwidth}{\labelnumberwidth}%
 \setlength{\leftmargin}{0pt}
 \setlength{\biblabelsep}{\labelsep}%
 \setlength{\itemsep}{\bibitemsep}%
 \setlength{\parsep}{\medskipamount}}%
 }
 {\endlist}
 {\item}
\DeclareNameAlias{default}{family-given}
\usepackage{markdown-jucs2e}

\DeclareMathOperator*{\argmin}{arg\,min}
\newcommand*\tran{^{\mkern-1.5mu\mathsf{T}}}
\newcommand\lmeta[1]{$\langle$\textit{#1}}
\newcommand\rmeta[1]{\textit{#1}$\rangle$}
\newcommand\meta[1]{$\langle$\textit{#1}$\rangle$}
\newcommand\E{\mathbb{E}}
\newcommand\Var{\mathbb{V}\mathrm{ar}}
\renewcommand\vec[1]{\boldsymbol{#1}}

\makeatletter
\def\fps@figure{tb}
\def\fps@table{b}
\makeatother

\begin{document}
\title{{\fontsize{14pt}{14pt}\selectfont{\vspace*{-3mm}\longpapertitle\vspace*{-1mm}}}}

\author{
   {\bfseries\fontsize{10pt}{10pt}\selectfont{Vít Novotný}} \\
   {\fontsize{9pt}{12pt}\selectfont{(Faculty of Informatics Masaryk University, Brno, Czech Republic\\
   \orcid{0000-0002-3303-4130}, 
   witiko@mail.muni.cz)}}
   \and
   {\bfseries\fontsize{10pt}{10pt}\selectfont{Michal Štefánik}} \\
   {\fontsize{9pt}{12pt}\selectfont{(Faculty of Informatics Masaryk University, Brno, Czech Republic\\
   \orcid{0000-0003-1766-5538}, 
   stefanik.m@mail.muni.cz)}}
   \and
   {\bfseries\fontsize{10pt}{10pt}\selectfont{Eniafe Festus Ayetiran}} \\
   {\fontsize{9pt}{12pt}\selectfont{(Department of Mathematical Sciences, Achievers University, Owo, Nigeria\\Faculty of Informatics Masaryk University, Brno, Czech Republic\\
   \orcid{0000-0002-6816-2781}, 
   ayetiran@mail.muni.cz)}}
   \and
   {\bfseries\fontsize{10pt}{10pt}\selectfont{Petr Sojka}} \\
   {\fontsize{9pt}{12pt}\selectfont{(Faculty of Informatics Masaryk University, Brno, Czech Republic\\
   \orcid{0000-0002-5768-4007}, 
   sojka@fi.muni.cz)}}
   \and
   {\bfseries\fontsize{10pt}{10pt}\selectfont{Radim Řehůřek}} \\
   {\fontsize{9pt}{12pt}\selectfont{(RARE Technologies Ltd.,
   radim@rare-technologies.com)}}
}

\maketitle

{\fontfamily{ptm}\selectfont
\begin{abstract}
{\fontsize{9pt}{9pt}\selectfont{\vspace*{-2mm}
\begin{markdown}

<!-- Objective -->
In 2018, @mikolov2018advances introduced the positional language model, which has
characteristics of attention-based neural machine translation models and which achieved
state-of-the-art performance on the intrinsic word analogy task.  However, the positional model
is not practically fast and it has never been evaluated on qualitative criteria or extrinsic tasks.
<!-- Methods  -->
We propose a *constrained positional model*, which adapts the sparse attention mechanism from neural machine translation to improve the speed of the positional model. We evaluate the positional
and constrained positional models on three novel qualitative criteria and on language modeling.
<!-- Results    -->
We show that the positional and constrained positional models contain
interpretable information about the grammatical properties of words and outperform other shallow models on language modeling. We also show that
our constrained model outperforms the positional model on language modeling and trains twice as fast.

 [^colab]: To make it easy for others to reproduce and build upon our work, we have published our experimental code at \url{https://github.com/MIR-MU/pine/\\#tutorials}.

\end{markdown}
}}
\end{abstract}}

{\fontfamily{ptm}\selectfont
}

{\fontfamily{ptm}\selectfont
\begin{category}
{\fontsize{9pt}{9pt}\selectfont{
G.1.3, G.3, I.2, I.2.7}}
\end{category}}

{\fontfamily{ptm}\selectfont
\begin{doi}
{\fontsize{9pt}{9pt}\selectfont{10.3897/jucs.69619}}
\end{doi}}

\vspace*{0.5cm}
\leavevmode
\noindent\null\hfill ``Words do not mean, people do.''~\cite{wright2008}

\vspace*{-2.75em}

\begin{markdown}

# Introduction
\label{sec:introduction}
Word representations of shallow log-bilinear language models (**LBL**s) such as word2vec [@mikolov2013efficient; @mikolov2013distributed] and fastText [@bojanowski2017enriching] have
found many applications in natural language processing (**NLP**) including word
similarity, word analogy, and language modeling [@bojanowski2017enriching; @ayetiran2021eds] as
well as word sense disambiguation [@chen2014unified; @ayetiran2021eds], text classification
[@kusner2015from; @novotny2020text], semantic text similarity [@charlet2017simbow], and
information retrieval [@novotny2020three, Section 4].
Recently, @devlin2018bert introduced the deep attention-based language
model **BERT**, which has redefined the state of the art for eleven **NLP**
tasks and turned **LBL**s to a baseline. Independently, @mikolov2018advances
have introduced the positional **LBL**, which resembles attention-based language
models and which has reached state-of-the-art performance on the intrinsic
word analogy task.\looseness=-1

@clark2019does also showed that ensembling **LBL**s with **BERT**
improves performance on the dependency parsing task compared to either **LBL**s
or **BERT** alone, which has reinvigorated the fading interest in **LBL**s.
Although surprising, the results of @clark2019does are supported by cognitive
psychology: @kahneman2011 describes the human mind as an interplay of two systems: the fast,
intuitive, and emotional System 1, and the slow, effortful and logical System
2. @peters2006numeracy have shown that systems 1 and 2 are mutually supportive
and that System 1 adapts to new and more challenging tasks by coaching System 2
to take over its more menial tasks. If we treat Kahnemann's
systems 1 and 2 as a metaphor for **LBL**s and **BERT**, the results of
@clark2019does seems natural.

In our paper, we describe the relationship between the attention mechanism and
the positional **LBL** of @mikolov2018advances, and we propose our constrained
positional **LBL** that adapts the attention sparsification techniques of
@dai2019transformerxl [; @child2019generating; @beltagy2020longformer;
@zaheer2020bigbird] for **LBL**s. We also develop three novel qualitative criteria,
which we use to evaluate the positional and constrained positional **LBL**s in
addition to the extrinsic language modeling task.

The rest of our paper is structured as follows: In
Section~\ref{sec:models}, we describe the dense and sparse attention mechanisms,
and we relate them to the positional **LBL** of @mikolov2018advances and our
proposed constrained positional **LBL**. In
Section~\ref{sec:experimental-setup}, we describe our experimental setup and
propose three novel qualitative evaluation measures.  In
Section~\ref{sec:results}, we discuss the results of our experiments.
We conclude in Section~\ref{sec:conclusion} by summarizing our results and
suggesting directions for future work.
\looseness=-1

# Models
\label{sec:models}
In this section, we describe the dense attention mechanism and we relate it to
the positional **LBL**. Additionally, we describe attention sparsification
techniques and we use them to develop our proposed constrained positional
**LBL**.

## Attention
\label{sec:attention}
In this section, we describe the purpose of attention in neural machine translation (**NMT**), and we describe sparsification techniques that make attention computationally tractable.

### Dense attention
\label{sec:dense-attention}
Early neural machine translation used *encoder-decoder models*, where an encoder recurrent neural network (**RNN**) would first read and encode a *source sequence* into a *fixed-length context vector* and a decoder **RNN** would then produce a *translated target sequence* from the context vector [@sutskever2014sequence; @cho2014learning]. Due to the context vector's fixed length, translation performance would deteriorate for longer sequences [@cho2014properties]. To enable the translation of longer source sequences, @bahdanau2014neural equipped the decoder with a dense *attention* mechanism. Instead of having a single encoded vector for the entire source sequence, the dense attention would construct a different context vector for each target word. Here, the context vector would be a weighted average of the encoder’s hidden states, where the weights would be trained to relate relevant source words to the target word.

Following the success of dense attention in **NMT**, @cheng2016long proposed to use the dense attention mechanism directly in the long-short-term memory (**LSTM**) cells of **RNN**s: Instead of computing the current *memory* and *hidden state* using the previous memory and hidden state alone, the current memory and hidden state would be computed as weighted averages of all previous memories and hidden states. Dense attention would act as a random-access memory mechanism, enabling the **LSTM** to recall long-range memories. Later, @vaswani2017attention proposed the *Transformer* architecture, which successfully replaced recurrence by the vertical stacking of dense attention, and which has been shown to be a *Turing-complete* [@perez2019turing] *universal approximator* [@yun2020transformers].

### Sparse attention
\label{sec:sparse-attention}
Since the dense attention mechanism learns weights for all pairs of source and target words, its space complexity is $\mathcal{O}(n^2)$ in the source sequence length. Several *sparse attention* architectures have been proposed in literature to enable the translation of longer source sequences by making the space complexity $\mathcal{O}(n)$.

@child2019generating proposed the *Sparse Transformer* architecture, which factorized the dense attention using $p$ separate attention heads to learn only $\mathcal{O}(n\cdot \sqrt[p]{n})$ weights. They showed that the resulting model could use larger context sizes and achieved significantly better results than Transformers on density modeling tasks.

Following the success of Sparse Transformers, @beltagy2020longformer proposed the *Longformer* architecture, which reduced the number of attention weights to $\mathcal{O}(n)$ and achieved significantly better results than Transformers on multiple long document tasks including question answering, coreference resolution, and
classification.

Finally, @zaheer2020bigbird proposed the *BigBird* architecture. Like Longformers, BigBird also used $\mathcal{O}(n)$ weights. Unlike Longformers, BigBird has been shown to be a Turing-complete universal approximator. Although attention enabled the recollection of long-range memories, sparse attention made it computationally tractable to do so.

## Log-bilinear language models
\label{sec:log-bilinear-language-models}
\label{sec:word2vec-and-fasttext}
In this section, we propose our constrained positional model that learns word representations while taking into account morphology and the mutual positions of words. When modeling positions of words, we only use a sparse subset of word vector features, following the hypothesis of @bach2012context that only a fraction of a word's meaning depends on the narrow context of a paragraph, whereas the rest of its meaning is either fixed or depends on a broader context.

We first present the general word2vec model of @mikolov2013efficient and @mikolov2013distributed, followed by the subword fastText model of @bojanowski2017enriching, and the positional model of @mikolov2018advances. Finally, we propose our constrained positional model together with its theoretical foundations, computational benefits, and its close relation to the sparse attention mechanism described in Section~\ref{sec:sparse-attention}.

### General model
\label{sec:general-model}
\label{sec:word2vec}
@mikolov2013efficient introduced the continuous bag of words (**CBOW**) model, which learns word representations by predicting a masked word $w_t$ from its context $C_t=w_{t-c}, \ldots, w_{t-1}, w_{t+1}, \ldots, w_{t+c}$, where $c$ is *window size* and $w_1, \ldots, w_T$ is the *training corpus*:
\end{markdown}
\begin{equation}
    \label{eq:optimization-problem}
    \argmin_{\vec{\theta}} \Big[L(\vec{\theta}) = -\sum_{t = 1}^T \log \Pr(w_t\mid C_t; \vec{\theta})\Big].\!
\end{equation}
\begin{markdown}
To estimate $\Pr(w_t\mid C_t)$, @mikolov2013distributed used a simplified variant of the noise contrastive approximation [@gutmann2012noise], which they called *negative sampling*:
\end{markdown}
\begin{equation}
    \label{eq:conditional-probability}
    \Pr(w_t\!\mid\!C_t)\!=\!\sigma(s(w_t, C_t))
        \!\!\!\!\prod_{n\in N_{C_t}}\!\!\!\!\sigma(-s(n, C_t)),\!
\end{equation}
\begin{markdown}
where $\sigma$ is the logistic function $x\mapsto \nicefrac{1}{1 + e^{-x}},$ $N_{C_t}$ is a set of negative examples $n$ for context $C_t$, and $s(w_t, C_t)$ is a scoring function that measures how well the masked word $w_t$ matches the context $C_t$:
\end{markdown}
\begin{equation}
    \label{eq:scoring-function-s}
    s(w_t, C_t) = \vec{u}_{C_t}\tran\cdot\vec{v}_{w_t},
    \vec{u}_{C_t} = \frac{1}{|C_t|}\!\sum_{w\in C_t}\!\!\vec{u}_{w}.
\end{equation}
\begin{markdown}
Here, $\vec{u}_{w}\in\mathbb{R}^D$ is the *input vector* of the context word $w$, $\vec{v}_{w_t}\in\mathbb{R}^D$ is the *output vector* of the masked word $w_t$, $\vec{u}_{C_t}$ is the *context vector* and $D$ is the number of word vector features, which is usually in the low hundreds.

@li1992random has shown that if we order words $w_{(i)}$ by their decreasing relative frequencies $f_{w_{(i)}}$ in a corpus, then $f_{w_{(i)}}$ exhibits a power law:
\begin{equation}
    f_{w_{(i)}} = \frac{c}{i^\alpha}\text{, where }c\approx0.1\text{ and }\alpha\approx 1.
\end{equation}
This law, originally proposed for English by @zipf1932selective, shows that most words in our training corpus will only represent a small subset of our vocabulary. By the end of the training, **CBOW** will have overfit the input and output vectors of the few most frequent words, whereas it will have underfit the input and output vectors of most other words.
\looseness=-1

To equalize the number of training samples for vocabulary words, @mikolov2013distributed discard corpus words with the following probability:
\begin{equation}
    \textstyle \Pr_\text{discard}(w_t) = \max\Big(0, 1 - \sqrt{\frac{r}{f_{w_t}}}\Big),
\end{equation}
where the low-pass threshold $r$ ensures that rare words $w_t$ with $f_{w_t}\leq r$ are never discarded.

### Subword model
\label{sec:subword-model}
\label{sec:fasttext}
The **CBOW** model only learns representations for words that are present in the training corpus. Additionally, vectors for different inflectional forms of a word share no weights, which delays training convergence for morphologically rich languages.

In response, @bojanowski2017enriching have extended **CBOW** by modeling subwords instead of words: The input vector $\vec{u}_w$ for a word $w$ become a sum of the input vectors $\vec{u}_g$ for the subwords $g\in G_w$ of $w$:
\begin{equation}
    \vec{u}_w = \sum_{g\in G_w} \vec{u}_g.
\end{equation}

### Positional model
\label{sec:positional-model}
\begin{figure}
\centering
\includegraphics{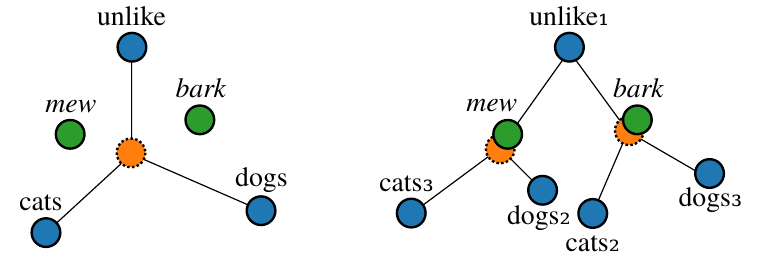}
\vspace\*{-1em}
\caption{The subword (left) and positional (right) models with their input (blue), context (orange), and output (green) vectors for the sentences “Unlike dogs, cats \meta{masked word}.” and “Unlike cats, dogs \meta{masked word}.” with two masked words: *mew* and *bark*.}
\label{fig:positional-model}
\end{figure}
In many sentences, the position of context words influences their syntactic function, which is important for predicting the masked word. Consider the following two sentences, which produce an identical context vector $\vec{u}_{C_t}$ despite the different masked words:

1. Unlike dogs, cats \meta{masked word}.
2. Unlike cats, dogs \meta{masked word}.

\noindent
If the context $C_t$ is large, distant words only introduce noise to the context vector $\vec{u}_{C_t}$.
\looseness=-1

To better adapt to these situations, we would like to have separate input vectors $\vec{u}_{w, p}$ for different positions $p\in P$ of a context word $w$:
\begin{equation}
    \vec{u}_{C_t} = \frac{1}{|P|}\sum_{p\in P}\vec{u}_{w_{t + p}, p}.
\end{equation}
See also Figure~\ref{fig:positional-model}. Since this would increase the size of the vocabulary by a factor of $|P| = 2c$, @mikolov2018advances proposed the *positional weighting*:
\begin{equation}
    \vec{u}_{w_{t + p}, p} = \vec{u}_{w_{t + p}}\odot\vec{d}_p,
\end{equation}
Here, $\vec{d}_p \in \mathbb{R}^{D'}, p\in P$ are *positional vectors* with $D' = D$ features and $\odot:\mathbb{R}^D\times\mathbb{R}^D\to\mathbb{R}^D$ is the Hadamard vector product.

Technically, the positional model may seem different from dense attention: Positional vectors change the input vectors of context words, so that the vectors better reflect the relationship between the masked word and the context words given their relative positions. In contrast, dense attention relates each target word in the translated sequence to relevant context words in the source sentence. However, they both serve the same purpose: to make the context vector for a given masked/target word more meaningful.
\looseness=-1

Compared to the subword model, the positional model more than doubles the training time, since we need to compute the Hadamard product and for each gradient update of an input vector $\vec{u}_{w}\in\mathbb{R}^D$, we also need to update the weights of a positional vector $\vec{d}_p\in\mathbb{R}^{D}$. The model can benefit from larger contexts $C_t$, but the training time scales linearly with the context window size $c$, which makes the model even more expensive.

@mikolov2018advances used the positional model to improve the state-of-the-art accuracy on the English word analogy task by 5\%. This demonstrates the importance of relating different positions of a sentence when creating its representation.

### Constrained positional model
\label{sec:constrained-positional-model}
According to the hypothesis of @bach2012context, the meaning of most words is partially *fixed* and partially dependent on the *narrow context* of a paragraph as well as the *broader context* that includes the conversational setting, the time and location of an utterance, and salient common ground, which may or may not be captured in the text:

> To hold that certain terms are context sensitive is not to deny that they have dictionary meanings. The claim is not that their meanings vary with the context. It is that their (standing) meanings determine their contents as a function of contexts of their use. After all, we wouldn’t look words up in the dictionary if they didn’t have (fairly) stable meanings.
> [...]
> Indeed, two different kinds of context are involved. Narrow context consists of matters of objective fact to which the determination of the semantic contents of certain expressions are sensitive. Broad context is the conversational setting, the mutual cognitive context or salient common ground. It includes the current state of the conversation (what has just been said, what has just been referred to, etc.), the physical situation (if the parties are face to face), salient personal knowledge, and relevant broader common knowledge.

\noindent
For example, consider the following sentence:

- Fruit flies like \meta{masked word}.

\noindent
The sentence admits at least two interpretations:

1. \textls[-25]{what a fly likes (adj-noun-verb-\meta{mask}),}
2. \textls[-25]{how fruit flies (noun-verb-prep-\meta{mask}).}

\noindent
Some masked words, such as “moisture”, satisfy only the first interpretation. Others, such as “a vegetable”, satisfy both interpretations.

Let us now rearrange the sentence as follows:

- \meta{Masked word} flies like fruit.

\noindent
The rearranged sentence only admits the second interpretation. The masked words still include “a vegetable” but no longer “moisture”.

To better adapt to these situations, context vectors $\vec{u}_{C_t}$ should contain two types of features:

\markdownColumns=1

1. $D'$ narrow-context-dependent features that take the positions of context words into account and inhibit the prediction of *moisture* in the rearranged sentence, and
2. $D - D'$ fixed and broader-context-dependent features that disregard the positions of context words and encourage the prediction of “a vegetable” in both sentences.

\noindent
Since **CBOW** does not model the broader context, we cannot distinguish between fixed and broader-context-dependent features. However, we can reduce broader-context-dependence with diachronic **CBOW** [@yao2018dynamic].

In the positional model, $D = D'$. Therefore, no word vector features are either fixed or broader-context-dependent. To represent the parts of a word's meaning that are fixed or dependent on the broader context, we propose to constrain the number of positionally-dependent features as follows:
\begin{equation}
    0 < D'\ll D.
\end{equation}
We define the constrained Hadamard vector product $\odot\colon\mathbb{R}^{D}\times\mathbb{R}^{D'}\to\mathbb{R}^{D}, D' < D$:
\begin{equation}
    \vec{u}_{w_{t + p}}\odot\vec{d}_p = \vec{u}_{w_{t + p}}\odot
    [\vec{d}_p \overbrace{\begin{matrix}1 & 1 & & \ldots & & 1\end{matrix}}^{D - D'\text{ times}}].
\end{equation}

When $D'$ is small, the constrained positional model can reach the speed of the subword model while modeling both the fixed and the context-dependent parts of a word's meaning. The model can benefit from larger contexts $C_t$ without making the computational complexity of training impractical, which is also the purpose of the sparse attention mechanism.

# Experimental setup
\label{sec:experimental-setup}
In this section, we describe our baseline, the initialization of weights, the hyperparameter and parameter optimization, the qualitative evaluation measures, the extrinsic **NLP** tasks used for performance estimation, and our training corpora.

## Baseline
\label{sec:baseline}
In our experiments, we compare our constrained positional model against the subword and positional models described in Section~\ref{sec:word2vec-and-fasttext}. For the subword model, we use the implementation in Gensim 3.8.3 [@rehurek2010software]. Since no public implementation of the positional model exists, we release our own implementation [as a free open-source software library.](https://github.com/MIR-MU/pine)

## Initialization
\label{sec:initialization}
For the general model, we follow the implementation of @bojanowski2017enriching and we initialize the features $u_i$ of the input word vectors $\vec{u}_{w}$ as i.i.d.\ r.v.'s with continuous uniform distribution:
\end{markdown}
\begin{equation}
    \label{eq:initialization-of-input-word-vectors}
    \vec{u}_{w} = (u_1, \ldots, u_D),
    u_i \sim \mathcal U\Big(\!\pm\frac{1}{D}\Big).
\end{equation}
\begin{markdown}
We initialize the output word vectors $\vec{v}_{w_t}$ to zero.
For the subword model, we initialize the input subword vectors $\vec{u}_g$ as in (\ref{eq:initialization-of-input-word-vectors}) and we also initialize the output subword vectors $\vec{v}_g$ to zero.

For the positional model, @mikolov2018advances do not describe the initialization of either the input subword vectors $\vec{u}_g$ or the positional vectors $\vec{d}_p$. Since no public implementation exists either, we initialize the features $u_i$ of $\vec{u}_g$ and the features $d_j$ of $\vec{d}_p$ as i.i.d.\ r.v.'s with the square-root normal distribution $\mathcal{N}^{0.5}(\mu, \sigma^2)$ of @pinelis2018exp:
\end{markdown}
\begin{equation}
\begin{split}
    \label{eq:initialization-of-positional-vectors}
    \vec{u}_{g} = (u_1, \ldots, u_{D}), 
    \vec{d}_{p} = (d_1, \ldots, d_{D'}), \\
    u_i \sim d_j \sim \mathcal{N}^{0.5}(\mu, \sigma^2),
    \mu = 0, \sigma^2 = \frac{1}{3D^2}.
\end{split}
\end{equation}
\begin{markdown}
\noindent
See also Appendix~\ref{app:initialization-of-the-positional-model}, where we show two other initialization options for the positional model, discuss their properties, and show their practical effect on the training of the model.
\looseness=-1

For the constrained positional model, we initialize the first $D'$ features of $\vec{u}_g$ and $\vec{d}_p$ as in \eqref{eq:initialization-of-positional-vectors} and the other $D - D'$ features of $\vec{u}_g$ and $\vec{d}_p$ as in \eqref{eq:initialization-of-input-word-vectors}.

## Optimization
\label{sec:optimization}
In this section, we describe which hyperparameters of the subword, positional, and constrained positional models we set according to previous work and which hyperparameters we optimized using the English word analogy task. We also describe how we train the model parameters $\vec{\theta}$.

### Hyperparameters
\label{sec:hyperparameters}
For the subword, positional, and constrained positional models, we use the following hyperparameter values of @mikolov2018advances, which give state-of-the-art performance on the English word analogy task: We store subwords of size 3--6 in a vocabulary backed by a hash table with bucket size $2\cdot 10^6$. We discard words with less than $5$ occurrences in the corpus and we equalize the number of training samples with the low-pass threshold $r = 10^{-5}$. We use $D = 300$ features in the input and output subword vectors. For the negative sampling loss, we use $|N_{C_t}| = 10$ negative samples. For the backpropagation of the loss function $L$, we use the initial learning rate $\gamma_0 = 0.05$.

For the subword, positional, and constrained positional models, we optimize the context window size $c$, because unlike the subword model, the positional model should benefit from larger contexts. For the constrained positional model, we optimize the number of positional features $D'$ to find the proper ratio between the fixed, narrow-context-dependent, and broader-context-dependent parts of a word's meaning.
To find the optimal hyperparameter values, we maximize a model's accuracy on the English word analogy task [@mikolov2013distributed] using Sequential Model-Based Optimization with the Tree-structured Parzen Estimator. Like @grave2018learning, we restrict the vocabulary for word analogies to the $2\cdot 10^5$ most frequent words in the training corpus.
\looseness=-1

### Parameters
\label{sec:parameters}
Following @bojanowski2017enriching, we optimize the model's parameters $\vec{\theta}$ by stochastic gradient descent over one epoch with the loss function $L(\vec{\theta})$ presented in \eqref{eq:optimization-problem} and with a linear decay of the learning rate $\gamma_t$ from $\gamma_0$ to zero:
\looseness=-1
\begin{equation}
    \gamma_t = \gamma_0\cdot \Big(1 - \frac{t}{T}\Big).
\end{equation}
We optimize the parameters in parallel using the HogWild lock-free approach of @recht2011hogwild with 8~Intel Xeon X7560 2.26\,GHz **CPU** cores. Since the optimization problem \eqref{eq:optimization-problem} is not sparse w.r.t.\ the positional vectors $\vec{d}_p$, most of which are updated at each training step, HogWild is less appropriate for the positional and constrained positional models than for the general and subword models. For all models, we report training times.

## Qualitative evaluation
\label{sec:qualitative-evaluation}
In this section, we propose qualitative evaluation measures, which we use to show the properties of the positional and constrained positional models. See also Appendix~\ref{app:qualitative-evaluation-measures}, where we show how our proposed measures relate to the conditional probability $\Pr(w_t \mid C_t)$ from~\eqref{eq:conditional-probability}.

### Masked word prediction
\label{sec:masked-word-prediction}
For the example sentences $C_t$ of the positional and constrained positional models from Section~\ref{sec:log-bilinear-language-models}, we show masked words $w_t$ in the descending order of the conditional probabilities $\Pr(w_t \mid C_t)$ from~\eqref{eq:conditional-probability}.

### Importance of positions
\label{sec:importance-of-positions}
For each position $p$, we show the min-max-scaled $\ell_2$-norm $\Vert\vec{d}_p\Vert$ of the positional vector $\vec{d}_p$, which measures the importance of position $p$ for predicting masked words.

Additionally, we cluster the $D'$ features $d_{p,j}$ of the positional vectors $\vec{d}_p$. For each cluster~$J$ and a position~$p$, we show the mean absolute value $\nicefrac{1}{|J|}\cdot\sum_{j\in J} |d_{p,j}|$, which measures the importance of position $p$ according to cluster~$J$.

### Importance of context words
\label{sec:importance-of-context-words}
For clusters $J$ and context words $w$, we use the mean absolute value $\nicefrac{1}{|J|}\cdot\sum_{j\in J} |u_{w,j}|$ to measure the importance of context words $w$ using cluster $J$, where $u_{w, j}$ are the features of the input vector $\vec{u}_{w}$ for $w$. For each cluster $J$, we show context words whose importance is maximized by~$J$.

## Performance estimation
\label{sec:performance-estimation}
In this section, we describe the extrinsic language modeling task, which we used to estimate the performance of the input word vectors $\vec{u}_{w}$ produced by the subword, positional, and constrained positional models.

### Language modeling
\label{sec:language-modeling}
For language modeling, we use a recurrent neural network (**RNN**) with the following architecture:

\markdownColumns=1

1. an input layer mapping a vocabulary $V$ of words $w$ to their *frozen* input vectors $\vec{u}_w$,
2. two hidden layers with $D=300$ **LSTM** units,
3. a fully-connected linear layer of size $|V|$, and
4. a softmax output layer that computes a probability distribution over the vocabulary $V$ using tied weights [@inan2017tying].

We evaluate our language model on [the English datasets][wmt13] introduced by @botha2014compositional and we report the validation and test perplexities. We use the same preprocessing and data splits as @botha2014compositional.

 [wmt13]: http://bothameister.github.io/

To train the **RNN**, we use stochastic gradient descent over 50 epochs, negative log-likelihood loss, dropout 0.5, batch size 40, and an initial learning rate 20 that is divided by 4 after each epoch with no decrease of validation loss. We clip gradients with $\ell_2$-norm above 0.25.

## Datasets
\label{sec:datasets}
For hyperparameter optimization, parameter optimization, qualitative evaluation, and performance estimation, we use the [2017 English Wikipedia dataset][enwiki] over a single epoch as our training corpus. The dataset contains 14.3\,GiB of raw text.

 [enwiki]: https://github.com/RaRe-Technologies/gensim-data (release wiki-english-20171001)

We preprocess our dataset by lower-casing and by tokenizing to longest sequences of Unicode characters with the *word* property. After tokenization, our dataset contains 2,423,655,228 words.

# Results
\label{sec:results}
In this section, we show and discuss the results of hyperparameter and parameter optimization, qualitative evaluation, and performance estimation.

## Optimization
\label{sec:word-analogy-results}

| Model                  | $c$ | $D'$ | Training Time          |
|:-----------------------|----:|-----:|:-----------------------|
| Subword                |   5 |      | 1 hour  and 11 minutes |
| Positional             |  15 |  300 | 4 hours and 12 minutes |
| Constrained positional |  15 |   60 | 2 hours and  5 minutes |
: The optimal context window sizes $c$ and numbers of positional features $D'$, and training times in hours for the subword, positional, and constrained positional models. \label{tab:optimal-hyperparameters-and-training-times}

Table~\ref{tab:optimal-hyperparameters-and-training-times} shows that the positional and constrained positional models benefit from larger contexts compared to the subword model. This is further evidenced by Figure~\ref{fig:window-size-to-accuracy}, which shows that the accuracy of the subword model steadily declines as the window size increases, whereas the positional model can cope with window sizes up to 40.

\begin{figure}
\vspace\*{1em}
\centering
\includegraphics{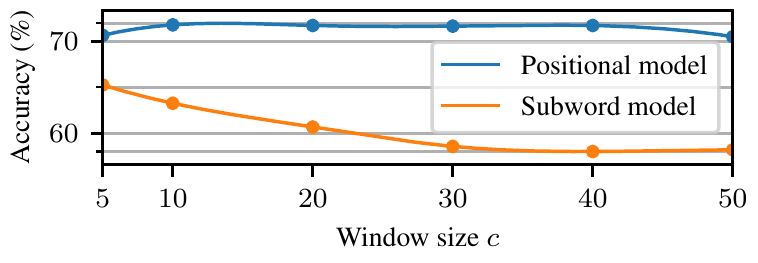}
\vspace\*{-1em}
\caption{Word analogy accuracy of the subword and positional models trained with different context window sizes $c$.}
\label{fig:window-size-to-accuracy}
\end{figure}

\begin{figure}
\vspace\*{1em}
\centering
\includegraphics{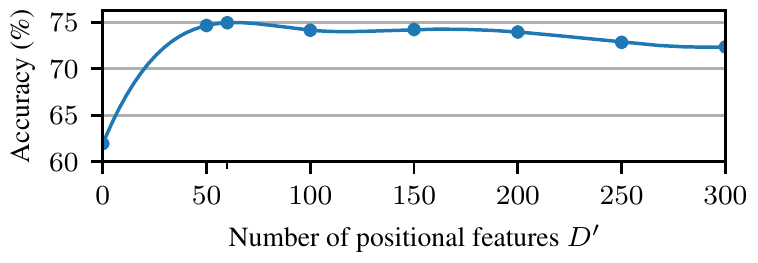}
\vspace\*{-1em}
\caption{Word analogy accuracy of the constrained positional model trained with different numbers of positional features $D'$.}
\label{fig:d-prime-to-accuracy}
\end{figure}

Table~\ref{tab:optimal-hyperparameters-and-training-times} also shows that the reduction of positional dimensionality $D'$ halves the training time of the constrained positional model compared to the positional model. Figure~\ref{fig:d-prime-to-accuracy} shows that the reduction of positional dimensionality also improves the accuracy of the constrained positional model compared to the positional model.

## Masked word prediction
\label{sec:masked-word-prediction-results}
Table~\ref{tab:masked-word-prediction} shows that the positional model predicts:
\end{markdown}
\begin{align}
  \Pr(\text{mew}\mid C_t^1) &> \Pr(\text{bark}\mid C_t^1), \\
  \Pr(\text{mew}\mid C_t^2) &< \Pr(\text{bark}\mid C_t^2).
\end{align}
\begin{markdown}
This matches our expectations and indicates that the model's context vectors contain narrow-context-dependent features that take the positions of context words “dogs” and “cats” into account.

Table~\ref{tab:masked-word-prediction} also shows that the constrained positional model predicts:
\end{markdown}
\begin{align}
  \Pr(\text{moisture}\mid C_t^3)  &>       \Pr(\text{moisture}\mid C_t^4), \\
  \Pr(\text{vegetable}\mid C_t^3) &\approx \Pr(\text{vegetable}\mid C_t^4).
\end{align}
\begin{markdown}
This indicates that the model contains not only narrow-context-dependent features that take the position of “moisture” into account, but also fixed and broader-context-dependent features that disregard the position of “a vegetable”.

## Importance of positions and context words
\label{sec:positional-vectors}
\label{sec:positional-vectors-results}
Figure~\ref{fig:position-importance} shows that in the positional and constrained positional models, the importance of positions $p\in [-2; 2]$ sharply decreases with their distance from the masked word. This shows that one of the basic functions of positional weighting is the attenuation of distant context words.
  
In the positional model, the importance of positions $p\not\in [-2; 2]$ increases with their distance from the masked word and even exceeds the importance of position $p=-2$ in the distant left context $p<-12$. In the constrained positional model, the importance of positions $p\not\in [-2; 2]$ is almost constant. Below, we will explain the cause of this difference using cluster analysis.

Figure~\ref{fig:clustered-importance} shows that in the positional model, the features of positional vectors fall into three main clusters: The two smaller clusters, which we call *antepositional* and *postpositional*, and the bigger cluster, which is missing from the constrained positional model and which we call *informational*.

\end{markdown}
\begin{table*}[t]
\leavevmode
\scalebox{0.975}{%
\small
\begin{tabular}{rl@{}l@{\enspace}rl}
\multicolumn{2}{c}{\small $C_t^1 = {}$“Unlike dogs,} & &
\multicolumn{2}{c}{\small $C_t^2 = {}$“Unlike cats,} \\
\multicolumn{2}{c}{\small cats \meta{masked word}.”} & &
\multicolumn{2}{c}{\small dogs \meta{masked word}.”} \\[0.5em] \cline{1-2}\cline{4-5}
      \# & Prediction  & &       \# & Prediction  \\ \cline{1-2}\cline{4-5}
       1 & cats        & &        1 & kennels     \\
       2 & spayed      & &        2 & cats        \\
       3 & kennels     & &        3 & puppies     \\
$\vdots$ & & &\multirow{2}{*}{$\vdots$} &         \\
    1820 & mew (100\%) & &          &             \\
  $\vdots$ &             & &   4065 & bark (99.9\%) \\
    5581 & bark (99.7\%)&& $\vdots$ &             \\
$\vdots$ & & &    5623 & mew (99.8\%) \\ \cline{1-2}\cline{4-5}
\multicolumn{5}{c}{\rule{0pt}{1.5em}\small (a) Positional model} \\
\end{tabular}%
\enspace
\begin{tabular}{rl@{}l@{\enspace}rl}
\multicolumn{2}{c}{\small $C_t^3 = {}$“Fruit flies} & &
\multicolumn{2}{c}{\small $C_t^4 = {}$“\lmeta{Masked}} \\
\multicolumn{2}{c}{\small like \meta{masked word}.”} & &
\multicolumn{2}{c}{\small \rmeta{word} flies like fruit.”} \\[0.5em] \cline{1-2}\cline{4-5}
      \# & Prediction  & &       \# & Prediction  \\ \cline{1-2}\cline{4-5}
       1 & fruit       & &        1 & fruit       \\
       2 & flies       & &        2 & insects     \\
       3 & insects     & &        3 & flies       \\
$\vdots$ & & &\multirow{2}{*}{$\vdots$} &         \\
     246 & vegetable (99.9\%)&&     &             \\
$\vdots$ &             & &      259 & vegetable (99.9\%) \\
    9036 & moisture (69.6\%)&&$\vdots$ &          \\
$\vdots$ & & &   33465 & moisture (42.8\%) \\ \cline{1-2}\cline{4-5}
\multicolumn{5}{c}{\rule{0pt}{1.5em}\small (b) Constrained positional model} \\
\end{tabular}%
}%
\vspace*{-0.5em}
\caption{Masked words $w_t$ predicted by the positional and constrained positional models for four example sentences. For selected words, we also show the conditional probability $P(w_t\mid C_t)$ in parentheses.}
\label{tab:masked-word-prediction}
\end{table*}
\begin{markdown}
\par

\begin{figure}
\vspace\*{1em}
\centering
\includegraphics{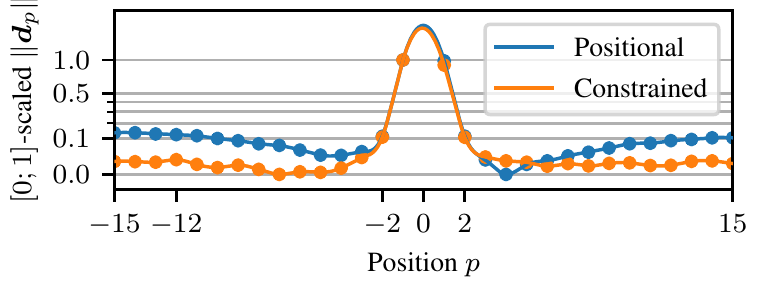}
\vspace\*{-1em}
\caption{The importance of different positions $p$ for predicting masked words in the positional and constrained positional models.}
\label{fig:position-importance}
\end{figure}

\begin{figure}
\centering
\includegraphics{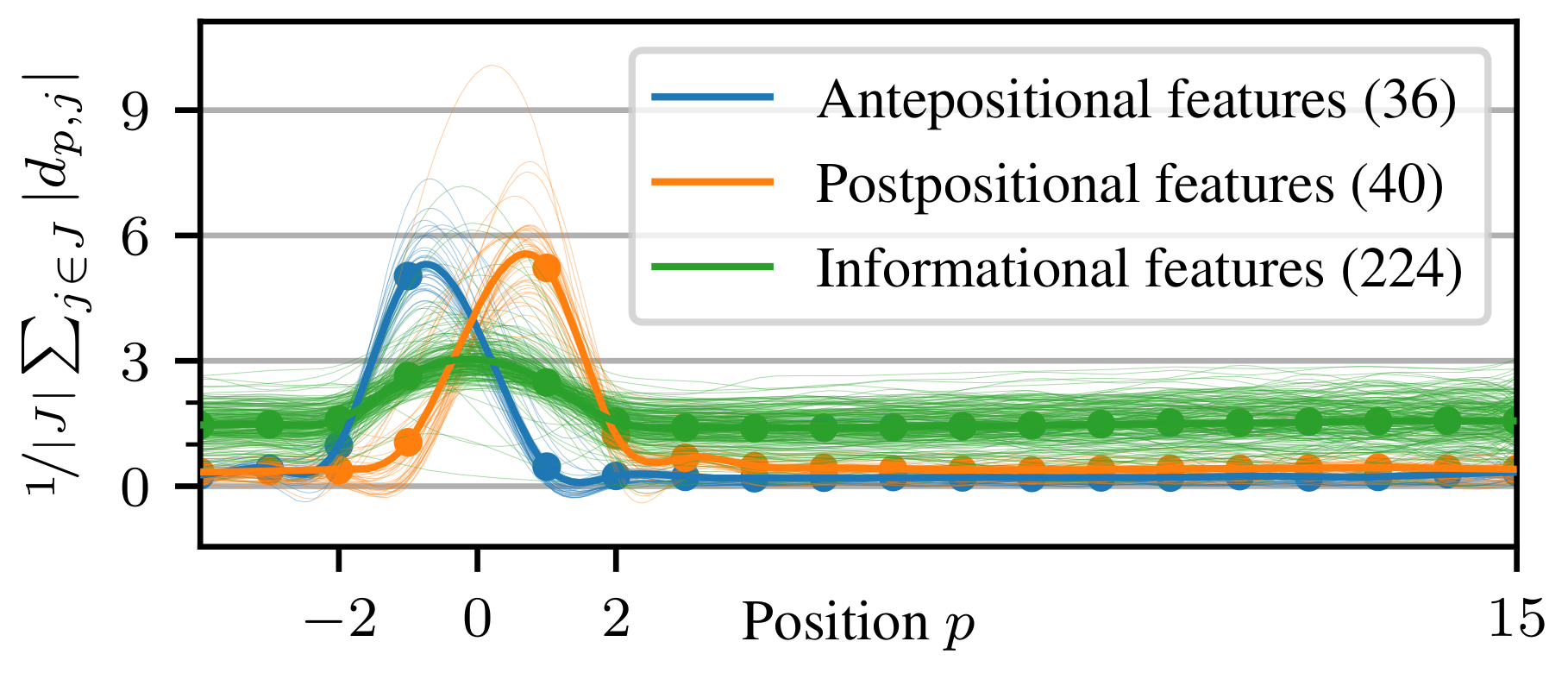}
\includegraphics{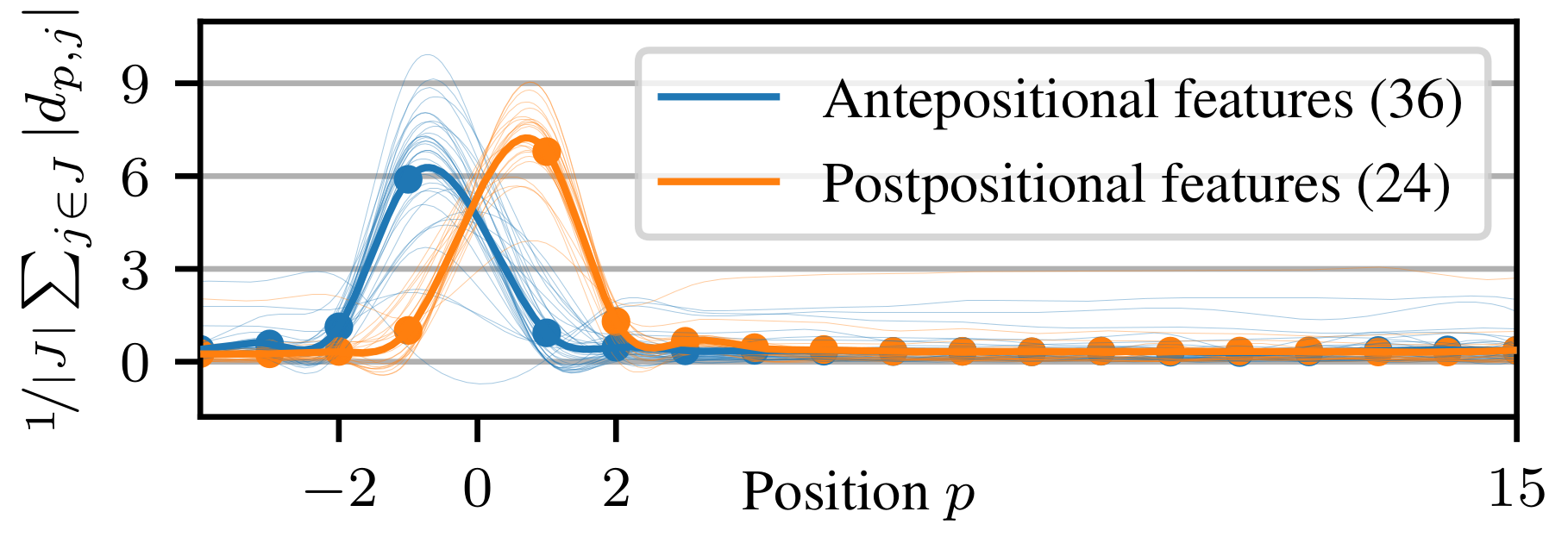}
\vspace\*{-1em}
\caption{The importance of different positions $p$ for predicting masked words in the positional (top) and constrained positional (bottom) models according to different clusters $J$ of positional features. For each cluster $J$, we show its size $|J|$ in parentheses.}
\label{fig:clustered-importance}
\end{figure}

The antepositional and postpositional features increase the importance of positions $p$ in anteposition ($-2, -1$) and in postposition ($1, 2$) of the masked word, respectively. Context words whose importance is maximized by antepositional features include “in”, “for”, and “coca”. Context words whose importance is maximized by postpositional features include “ago”, “else”, and “cola”. The number of antepositional and postpositional features in the positional model is 76, which is close to the $D'=60$ positional features selected for the constrained positional model by hyperparameter optimization. This indicates that the highest task performance is reached when only antepositional and postpositional features remain.

The informational features increase the importance of positions $p\not\in [-2; 2]$. @levy2014neural showed that context words with large input vectors have high self-information. We believe that the purpose of informational features is to amplify distant self-informational context words that indicate the general topic of a sentence. Context words whose importance is maximized by informational features include “finance”, “sports”, and “politics”. In the constrained positional model, informational features $d_{p, j}$ are effectively replaced by ones, which is close to what the positional model has learnt.
\looseness=-1

## Language modeling
\label{sec:language-modeling-results}
Figure~\ref{fig:language-modeling} shows that the positional and constrained positional models consistently outperform the subword model during the training of **RNN** language models. Figure~\ref{fig:language-modeling} also shows that **RNN** language models have converged and that training for more epochs would not have improved their perplexity.
  
Table~\ref{tab:language-modeling} shows that the constrained positional model produces word vectors that are better suited for initializing the lookup tables of **RNN** language models than the subword and positional models.

\begin{figure}
\vspace\*{1em}
\centering
\includegraphics{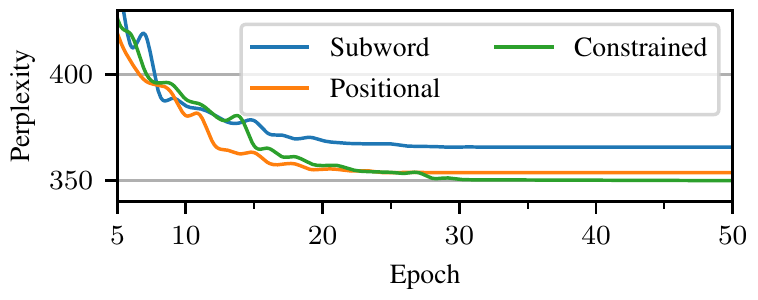}
\vspace\*{-1em}
\caption{Validation perplexities at different epochs of **RNN** language models that use subword, positional, and constrained positional models as their lookup tables.}
\label{fig:language-modeling}
\end{figure}

|| Subword | Positional | Constrained positional |
|-----------------|---------|-----------|--------|
| Test perplexity | 360.91  | 347.52    |*343.13*|
: Test perplexities of **RNN** language models that use subword, positional, and constrained positional models as lookup tables. Best result is *emphasized*. \label{tab:language-modeling}

# Conclusion and future work
\label{sec:conclusion}

In our work, we have related the attention mechanism from **NMT** to the positional language model of @mikolov2018advances and adapted the attention sparsification techniques
of @zaheer2020bigbird to develop our constrained positional model.

We have shown that our constrained positional model is $2\times$
faster to train and more accurate at modeling English than the positional model. Future work should focus at the quantitative evaluation on language modeling in languages other than English and on other extrinsic **NLP** tasks, both alone and together with deep language models.

Furthermore, we have developed three novel qualitative evaluation measures and we used them to show that the positional vectors in English positional and constrained positional models serves two distinct roles: They allow the use of larger context sizes and they determine the grammatical properties of words. Future work should investigate the role of positional vectors in languages other than English.

\end{markdown}

\begin{Acknowledgements}
The first author's work was graciously funded by the South Moravian Centre for International Mobility as a part of the Brno Ph.D.\ Talent project.
\end{Acknowledgements}

\begingroup
\let\emph\textit
\printbibliography
\endgroup

\appendix

\begin{markdown}

# Initialization of the positional model
\label{app:initialization}
\label{app:initialization-of-the-positional-model}
In this appendix, we expand on Section \ref{sec:initialization} by describing three initialization options for the positional model. We discuss the properties of the initialization options and their practical effect on the training of the model.

## Identity positions and vanilla subwords
\label{app:identity-positions-and-vanilla-subwords}
To keep the effective learning rate of the positional model the same as in the subword model, it is sufficient to keep the distribution of the context vector $\vec{u}_{C_t}$ the same as in the subword model. To achieve this, we initialize the input subword vectors $\vec{u}_g$ as in \eqref{eq:initialization-of-input-word-vectors} and the positional vectors $\vec{d}_p$ to one. Intuitively, the training starts with no positional weighting and the positional vectors are learnt later. In practice, $\vec{d}_p\gg\vec{u}_g$, causing the gradient $\nabla_{\vec{u}_g} L$ to explode for $D>600$ soon after the training has begun. This leads to numerical instability and the model parameters $\vec{\theta}$ tend to NaN as the training continues.\looseness=-1

## Positions same as vanilla subwords
\label{app:positions-same-as-vanilla-subwords}
The simplest option is to initialize both the input subword vectors $\vec{u}_g$ and the positional vectors $\vec{d}_p$ as in \eqref{eq:initialization-of-input-word-vectors}. In practice, this decreases the variance of the context vector $\vec{u}_{C_t}$ in the positional model compared to the subword model:
\end{markdown}
\begingroup
\newcommand\bigsum{\sum_{\substack{w\in C_t\\g\in G_{w}}}\!}
\begin{align}
    &\Var\Big[\frac{1}{|C_t|} \bigsum\! \vec{u}_g\odot\vec{d}_p\Big] =
    \frac{1}{|C_t|^2} \!\bigsum\! \Big(\Var[\vec{u}_g\odot\vec{d}_p] = \E\big[\vec{u}_g^2\big]\odot\E\big[\vec{d}_p^2\big]\Big) \\&=
    \frac{1}{|C_t|^2} \!\bigsum \vec{\frac{1}{9D^4}} \ll
    \frac{1}{|C_t|^2} \!\bigsum \vec{\frac{1}{3D^2}} =
    \frac{1}{|C_t|^2} \!\bigsum\! \Var[\vec{u}_g] =
    \Var\Big[\frac{1}{|C_t|} \sum_{\substack{w\in C_t\\g\in G_{w}}} \vec{u}_g\Big]. \notag
\end{align}
\endgroup
\begin{markdown}

\begin{figure\*}[p]
\centering
\includegraphics{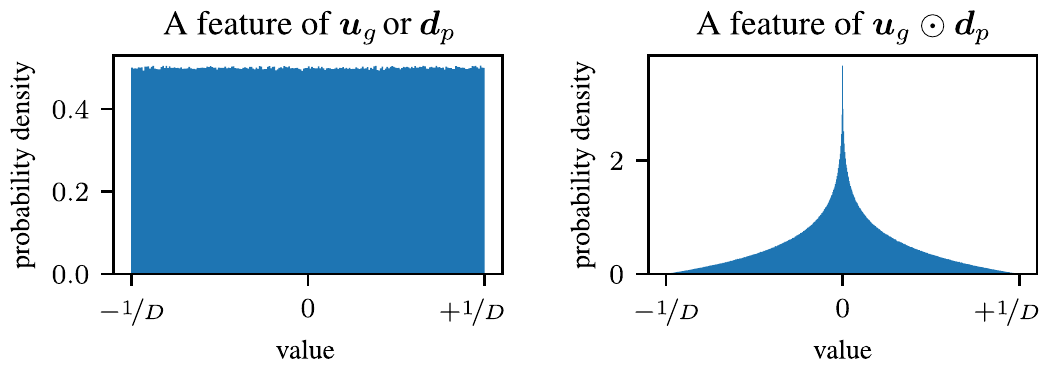}
\vspace\*{-1em}
\caption{Probability densities of feature values in the input subword vectors $\vec{u}_g$, the positional vectors $\vec{d}_p$, and their product $\vec{u}_g\odot\vec{d}_p$ with the *positions same as vanilla subwords* initialization to $\mathcal U(\pm\nicefrac{1}{D}), D=1$. Since $\Var[\vec{u}_g] \gg \Var[\vec{u}_g\odot\vec{d}_p]$, the effective learning rate of the positional model is smaller than in the subword model.}
\label{fig:positions-same-as-subwords}
\end{figure\*}

\begin{figure\*}[p]
\vspace\*{1.5em}
\centering
\includegraphics{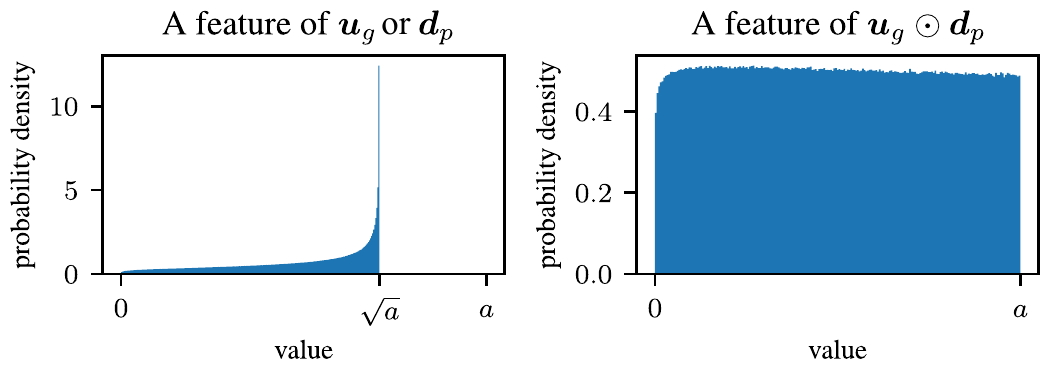}
\vspace\*{-1em}
\caption{Probability densities of feature values in the input subword vectors $\vec{u}_g$, the positional vectors $\vec{d}_p$, and their product $\vec{u}_g\odot\vec{d}_p$ with the initialization to $\mathcal U^{0.5}(0, a), a=2$. For our use, we would need $\mathcal U^{0.5}(\pm\nicefrac{1}{D})$ instead.}
\label{fig:positions-same-as-subwords-square-root-uniform}
\end{figure\*}

\begin{figure\*}[p]
\vspace\*{1.5em}
\centering
\includegraphics{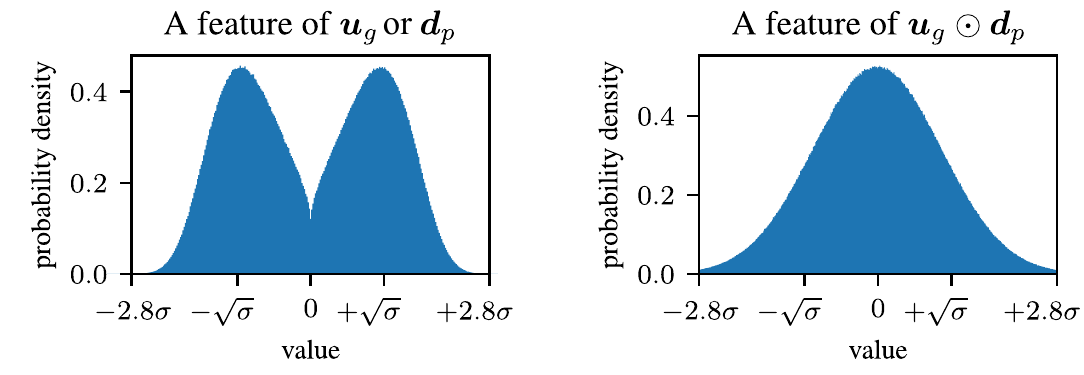}
\vspace\*{-1em}
\caption{Probability densities of feature values in the input subword vectors $\vec{u}_g$, the positional vectors $\vec{d}_p$, and their product $\vec{u}_g\odot\vec{d}_p$ with the *positions same as subwords* initialization to the square-root normal distribution $\mathcal{N}^{0.5}(0, \sigma^2),\sigma^2=\nicefrac{1}{3D^2}, D=1$.}
\label{fig:positions-same-as-subwords-square-root-normal}
\end{figure\*}

\begin{figure}[t]
\centering
\includegraphics{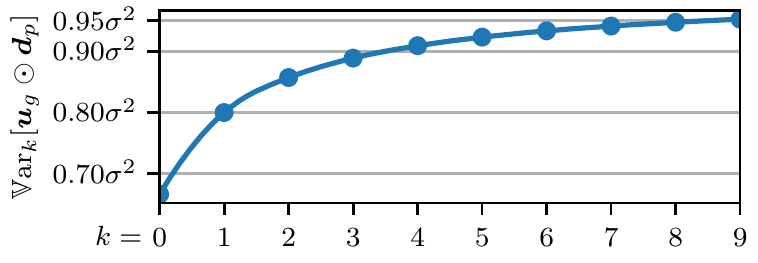}
\vspace\*{-1em}
\caption{The variances $\Var_k[\vec{u}_g\odot \vec{d}_p]$ for the feature values in the product $\vec{u}_g\odot \vec{d}_p$ of the input subword vectors $\vec{u}_g$ and the positional vectors $\vec{d}_p$ with the initialization to the square-root normal distribution $\mathcal{N}^{0.5}(0, \sigma^2)$, when we approximate the infinite sequence $(a_n)_{n=0}^\infty$ in the definition of $\mathcal{N}^{0.5}$ by first $k+1$ elements $(a_n)_{n=0}^k$.}
\label{fig:positions-same-as-subwords-variance}
\end{figure}

See also Figure~\ref{fig:positions-same-as-subwords}. The decrease in $\Var[\vec{u}_{C_t}]$ decreases $\Var[\nabla L]$ and therefore it also decreases the effective learning rate of the positional model. As $D$ increases, the context vector $\vec{u}_{C_t}$ quickly tends to zero due to $\Var[\vec{u}_g\odot\vec{d}_p] = \vec{\nicefrac{1}{9D^4}}$.
   
## Positions same as subwords
\label{app:positions-same-as-subwords}
To keep the effective learning rate of the positional model the same as in the subword model and to avoid exploding gradients, it is sufficient to keep the distribution of the context vector $\vec{u}_{C_t}$ the same as in the subword model and to initialize both the input subword vectors $\vec{u}_g$ and the positional vectors $\vec{d}_p$ from the same distribution. We could achieve this by initializing the features $u_i$ of $\vec{u}_g$ and the features $d_j$ of $\vec{d}_p$ as i.i.d.\ r.v.'s with the square-root distribution $\mathcal{U}^{0.5}(\pm\nicefrac{1}{D})$ such that $u_i\cdot d_j\sim \mathcal{U}(\pm\nicefrac{1}{D})$. Although an approximation of $\mathcal{U}^{0.5}(0, a)$ using the $\beta$-distribution is known [@ravshan2018factor], see Figure~\ref{fig:positions-same-as-subwords-square-root-uniform}, it does not extend to $\mathcal{U}^{0.5}(\pm\nicefrac{1}{D})$, so we need another approach.
\looseness=-1

Assuming the context $C_t$ is sufficiently large, then by the central limit theorem, the features of the context vector $\vec{u}_{C_t}$ in the subword model have the normal distribution $\mathcal{N}(\mu, \nicefrac{\sigma^2}{|C_t|})$, where $\mu\\!=\\!\E[\mathcal{U}(\pm \nicefrac{1}{D})]\\!=\\!0, \sigma^2\\!\\!=\\!\\!\Var[\mathcal{U}(\pm \nicefrac{1}{D})]\\!=\\!\nicefrac{1}{3D^2}$. To achieve the same distribution with the positional model, we initialize the features $u_i$ of $\vec{u}_g$ and the features $d_j$ of $\vec{d}_p$ as i.i.d.\ r.v.'s with some continuous distribution $\mathcal{X}$ such that $\E[u_i\cdot d_j] = \mu$ and $\Var[u_i\cdot d_j] = \sigma^2$. In our initialization, we use as $\mathcal{X}$ the square-root normal distribution $\mathcal{N}^{0.5}(\mu, \sigma^2)$ of @pinelis2018exp, see Figure~\ref{fig:positions-same-as-subwords-square-root-normal}. The continuous uniform $\mathcal{U}(\pm\nicefrac{\sqrt[4]{3}}{\sqrt{D}})$ is also an option.

The definition of $\mathcal{N}^{0.5}$ by @pinelis2018exp contains an infinite sequence $(a_n)_{n=0}^\infty$:
\end{markdown}
\begin{equation}
    \mathcal{N}^{0.5}(\mu, \sigma^2) =
    \epsilon \cdot e^{\sum_{n=0}^\infty a_n} \cdot \sqrt{\sigma} + \sqrt\mu,
    a_n = \frac{1}{4}\cdot\ln\Big(1 + \frac{1}{\max(1, n)}\Big) - \frac{G_n}{2n + 1},
\end{equation}
\begin{markdown}
where $(G_n)_{n=0}^\infty$ are i.i.d.\ r.v.'s with $\text{Gamma}(\nicefrac{1}{2}, 1)$ distribution and $\epsilon$ is a Rademacher r.v.\ independent of all $G_n$. Since $\lim_{n\to\infty} a_n = 0$, we can approximate $\sum_{n=0}^\infty a_n$ by its first $k+1$ elements, but we need guarantees about $\E[u_i\cdot d_j]$ and $\Var[u_i\cdot d_j]$.
We can see that $\E[u_i\cdot d_j] = \mu$ for any $k$:
\end{markdown}
\begin{multline}
    E[u_i] = \E[d_j]
    = \E[\epsilon]\cdot\E\big[e^{\sum_{n=0}^k a_n}\big]\cdot\E[\sqrt{\sigma}] + \E[\sqrt\mu]
    = 0\cdot\E\big[e^{\sum_{n=0}^k a_n}\big]\\\cdot\E[\sqrt{\sigma}] + \sqrt\mu = \sqrt\mu,
    \E[u_i\cdot d_j] = \E[u_i]\cdot\E[d_j] = (\sqrt\mu)^2 = \mu.
\end{multline}
\begin{markdown}
We will denote the variance for a given $k$ as $\Var_k$:
\end{markdown}
\begin{equation}
\begin{aligned}
    &  \Var_k[u_i] = \Var_k[d_j]
    = \Var\big[\epsilon\cdot e^{\sum_{n=0}^k a_n}\cdot\sqrt{\sigma} + \sqrt\mu\big]
    \\&= \Big(\prod_{n=0}^k\E[e^{a_n}]^2 + \Var\Big[\prod_{n=0}^k e^{a_n}\Big]\Big)\cdot\sigma, \\
    &  \Var_k[u_i\cdot d_j] = 2\mu\cdot\Var_k[u_i] + \Var_k[u_i]^2. \\
\end{aligned}
\end{equation}
\begin{markdown}

In our initialization, we approximate $\sum_{n=0}^\infty a_n$ by $\sum_{n=0}^k a_n, k=9$. As we show in Figure~\ref{fig:positions-same-as-subwords-variance}, this guarantees $\E[u_i\cdot d_j] = \mu, \nicefrac{\Var[u_i\cdot d_j]}{\sigma^2} \in (0.95; 1]$.

# Qualitative evaluation measures
\label{app:qualitative-evaluation-measures}
In this appendix, we expand on Section \ref{sec:qualitative-evaluation} by showing how the proposed qualitative evaluation measures relate to the conditional probability $\Pr(w_t\mid C_t)$ from \eqref{eq:conditional-probability}. For a fixed set of negative samples $N_{C_t}$, $\Pr(w_t\mid C_t)$ is a strictly increasing transformation of the scoring function $s(w_t, C_t)$ from \eqref{eq:scoring-function-s}. Without loss of generality, our proofs will focus on $s(w_t, C_t)$ rather than on $\Pr(w_t\mid C_t)$.

## Importance of positions
\label{app:importance-of-positions}
All else being constant, the $\ell_2$-norm $\Vert \vec{d}_p\Vert$ is an asymptotic upper bound on $|s(w_t, C_t)|$:
\end{markdown}
\begin{align}
&|s(w_t, C_t)|
= |\vec{u}_{C_t}\tran\cdot\vec{v}_{w_t}|
= \frac{1}{|P|}\cdot| (\vec{u}_{w_{t+p}}\odot\vec{d}_p)\tran\vec{v}_{w_t} + \ldots|  \\
&\leq \frac{1}{|P|}\cdot(\Vert \vec{u}_{w_{t+p}}\Vert \cdot\Vert \vec{d}_p\Vert \cdot\Vert \vec{v}_{w_t}\Vert  + \Vert  \ldots\Vert ), \notag |s(w_t, C_t)| \in \mathcal{O}(\Vert \vec{d}_p\Vert). \notag
\end{align}
\begin{markdown}
\par
All else being constant, $\sum_{j\in J} |d_{p,j}|$ is also an asymptotic upper bound on $|s(w_t, C_t)|$:
\end{markdown}
\begin{align}
& |s(w_t, C_t)|
= |\vec{u}_{C_t}\tran\cdot\vec{v}_{w_t}|
= \frac{1}{|P|}\cdot\Big|\sum_{j\in J} u_{w_{t+p},j}\cdot d_{p, j}\cdot v_{w_t,j} + \ldots\Big|
\\&\leq \frac{1}{|P|}\cdot\Big(\!\sum_{j\in J} |u_{w_{t+p},j}|\cdot|d_{p, j}|\cdot|v_{w_t,j}| + |\ldots|\Big), \notag
|s(w_t, C_t)| \in \mathcal{O}\Big(\!\sum_{j\in J} |d_{p,j}|\Big), \notag
\end{align}
\begin{markdown}
where $u_{w_{t+p},j}$ are features of the input vector $\vec{u}_{w_{t+p}}$ for the context word $w_{t+p}$ at position $p$ and $v_{w_t,j}$ are features of the output vector $\vec{v}_{w_t}$ for the masked word $w_t$.

## Importance of context words
\label{app:importance-of-context-words}
All else being constant, $\sum_{j\in J} |u_{w,j}|$ is an asymptotic upper bound on the expected absolute difference $\E|s(w_t, C_t) - s(w_t, C'_t)|$ for a fixed context word $w$ and random-valued masked words $w_t$ and contexts $C_t, C'_t$, where $w$ is at position $p_1$ in $C_t$ and at position $p_2$ in $C'_t$:
\end{markdown}
\begin{multline}
\unskip \E|s(w_t, C_t) - s(w_t, C'_t)| \\
= \frac{1}{|P|}\cdot\sum_{w_t, C_t, C'_t}\Big|\sum_{j\in J} u_{w, j}\cdot (d_{p_1, j} - d_{p_2, j})\cdot v_{w_t, j}
+ \ldots\Big| \cdot \Pr(w_t, C_t, C'_t) \hfill\\
\leq \frac{1}{|P|}\cdot\sum_{\substack{w_t, C_t, C'_t\\ j\in J}}|u_{w, j}|\cdot |(d_{p_1, j} - d_{p_2, j})\cdot v_{w_t, j}|
\cdot \Pr(w_t, C_t, C'_t) + |\ldots|\cdot \Pr(w_t, C_t, C'_t), \\
\E|s(w_t, C_t)\!-\!s(w_t, C'_t)| \in \mathcal{O}\Big(\!\sum_{j\in J} |u_{w,j}|\Big).\hfill
\end{multline}
\end{document}